\documentclass[letterpaper, 10 pt, conference]{ieeeconf}

\IEEEoverridecommandlockouts
\usepackage{graphicx}
\usepackage{epsfig}
\usepackage{amsmath,amssymb}

\usepackage{amsthm}

\usepackage{tcolorbox}
\tcbuselibrary{skins, breakable}

\definecolor{babypurple}{HTML}{F5F1FF}
\definecolor{improvegreen}{HTML}{00BF00}

\theoremstyle{remark}
\newtheorem{remark}{Remark}

\tcolorboxenvironment{remark}{
  enhanced, breakable,
  colback=babypurple,
  frame hidden,
  arc=3pt,
  left=10pt, right=10pt, top=6pt, bottom=6pt,
  before skip=8pt, after skip=8pt,
}

\usepackage{multirow}
\usepackage{longtable}
\usepackage{booktabs}
\let\labelindent\relax
\usepackage[inline]{enumitem}

\usepackage[caption=false,font=footnotesize]{subfig}

\usepackage{xcolor}
\usepackage{colortbl}
\usepackage{url}
\usepackage[font=small]{caption}
\usepackage[ruled,vlined]{algorithm2e}
\usepackage{tikz}
\usetikzlibrary{arrows.meta, positioning}

\usepackage{cite}
\usepackage[colorlinks=true,
    linkcolor=purple,
    urlcolor=blue,
    citecolor=blue]{hyperref}

\usepackage{pifont}

\usepackage{tgstyle}

\begin{document}

\title{\LARGE \bf
Adaptive Multi-Modal Out-of-Distribution Detection for Trajectory Prediction in Autonomous Vehicles
}

\author{Tongfei Guo$^{1}$ and Lili Su$^{1}$%
\thanks{*This work was supported in part by the NSF CAREER award under grant~2340482. $^{1}$Department of Electrical and Computer Engineering, Northeastern University, Boston, MA, USA.
        {\tt\small \{guo.t, l.su\}@northeastern.edu}}%
\thanks{\scriptsize\textcopyright~2026 IEEE. Personal use of this material is permitted. Permission from IEEE must be obtained for all other uses, in any current or future media, including reprinting/republishing this material for advertising or promotional purposes, creating new collective works, for resale or redistribution to servers or lists, or reuse of any copyrighted component of this work in other works.}%
}

\maketitle

\begin{abstract}
Trustworthy trajectory prediction grounds autonomous vehicle (AV) safety, yet deployed models
inevitably face out-of-distribution (OOD) scenes. Prior AV OOD detection targets perception,
but planners act on predicted futures rather than raw scenes, so erroneous forecasts can slip past
frame-level checks and corrupt control. We therefore tackle OOD detection at the
trajectory-prediction level.
Our analysis of real-world benchmarks reveals that prediction errors are often \textit{multi-modal}, exhibiting distinct low- and high-error modes that evolve with open-world driving context.
Observing this, we propose \textit{Mode-Aware CUSUM}, which explicitly models multiple error modes while retaining the efficiency and general compatibility of classical CUSUM.
By dynamically identifying the active error mode and adapting detection thresholds, our method enables robust monitoring across heterogeneous conditions. Experiments on large-scale trajectory benchmarks demonstrate consistent reductions in detection delay and false alarms.
\end{abstract}

\section{Introduction}
Trajectory prediction is a safety-critical component of autonomous vehicles (AVs) directly informing downstream planning and control. Despite
impressive benchmark performance, deployed prediction models face the \emph{deployment
gap}. Large-scale datasets, while broad, underrepresent rare ``tail'' scenes such as
aggressive lane merges, emergency vehicles, or unusual road geometries. Real-world uncertainty, such as irregular blockages, adverse weather, and sensor degradation, routinely shift the live input away from what the model was trained on. In these \emph{out-of-distribution} (OOD) scenes,
models tend to produce overconfident but hazardous forecasts, potentially misleading the ego
vehicle's decision-making at the worst possible moment.
Timely OOD detection is thus essential: early flagging lets the planner fall back safely before errors propagate into control.

Existing research on OOD detection in AVs has primarily focused on computer vision tasks such as object detection and semantic
segmentation~\cite{9564545, shoeb2025out}, with representative approaches
including post-hoc scoring and energy-based models~\cite{hendrycks2016baseline, liu2020energy}. These frame-wise approaches evaluate whether a scene is in-distribution (ID) or OOD based solely on the current frame, overlooking the spatiotemporal correlations among the trajectories of surrounding agents.
Uncertainty quantification methods such as deep ensembles and Monte Carlo dropout~\cite{gal2016dropout, nayak2022uncertainty} can exploit temporal signals, but they incur heavy computational cost and offer no formal guarantee on detection delay.
Sequential CUSUM-based detection~\cite{10.1145/3803009} addresses both concerns, yet it assumes a single, stationary error distribution and therefore cannot track the latent, shifting error modes that arise in real traffic.

\begin{figure}[t!]
    \centering
    \vspace{0.6em}
    \includegraphics[width=\linewidth]{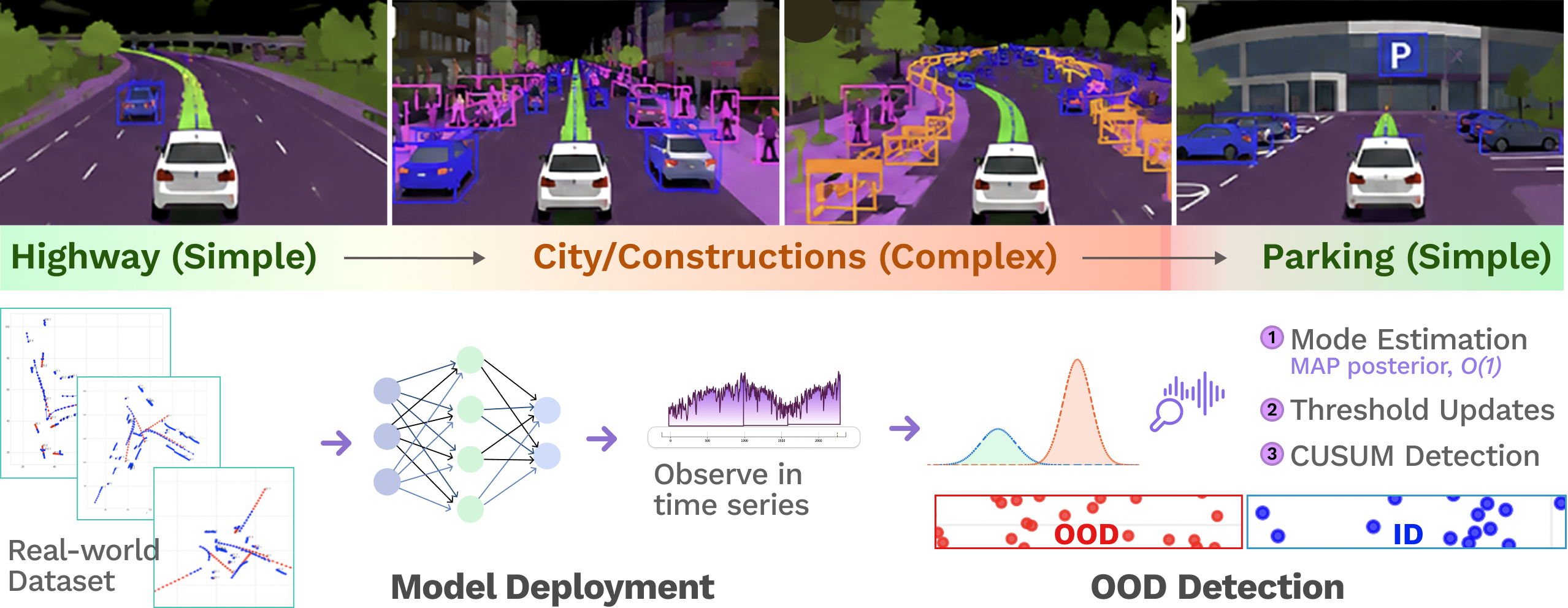}
    \caption{\textbf{Illustration of our adaptive multi-mode OOD detection framework.} {\it Top}: A representative driving sequence in which the ego vehicle traverses scenarios of varying difficulty ({\color{improvegreen} easy} $\rightarrow$ {\color{orange} complex} $\rightarrow$ {\color{improvegreen} easy}), where the surrounding agent types, agent speeds, scene complexity, and observation noise vary over time. {\it Bottom}: Architecture of the proposed framework, which monitors prediction errors in real time, dynamically identifies the active error mode (low/high), and adapts the detection threshold accordingly.}
    \label{fig:fig1}
\vspace{-15pt}
\end{figure}

Consider a vehicle traveling from a busy downtown to the airport. In the city center, it navigates dense stop-and-go traffic with frequent pedestrian and cyclist interactions, where prediction is difficult and errors run high. Once it merges onto the highway, the scene simplifies: traffic flows steadily, interactions thin out, and predictions become accurate. As the route unfolds, the complexity of interactions and road geometry shifts continuously, and the typical scale of prediction error shifts with it. A detector tied to one fixed threshold cannot serve both ends of this journey; the threshold must adapt, tracking how noisy the prediction error currently is.
Our empirical analysis on multiple real-world benchmarks~\cite{huang2018ApolloScape, fhwa2020, caesar2020nuscenes} confirms this intuition: prediction errors consistently exhibit a multi-modal distribution, and the active mode evolves over time (Fig.~\ref{fig:fig1} and Fig.~\ref{fig:GMM}, Sec.~\ref{sec: multiple modes}).

To address these limitations, we propose Mode-Aware CUSUM, an adaptive sequential detector that infers the latent error mode and applies a scene-adaptive threshold, preserving the $O(1)$ cost and formal guarantees of classical CUSUM while remaining robust across heterogeneous driving conditions.
Our framework requires only streaming prediction residuals---no model internals, ground-truth labels, or expensive uncertainty quantification. Moreover, we designed it agnostic to the specific models deployed, making it flexible and general.

\smallskip
\noindent\textbf{Contributions.} Our contributions can be summarized as:
\begin{itemize}[leftmargin=*]
\item We provide the first systematic evidence, across multiple widely-used trajectory prediction benchmarks, that prediction errors exhibit \emph{multi-modal} distributions, and the active mode evolves over time.

\item We introduce a \emph{Mode-Aware CUSUM} (MA-CUSUM) algorithm that adapts detection thresholds based on the estimated error mode.

\item We conduct extensive experiments on established benchmarks (ApolloScape, NGSIM, nuScenes), demonstrating that MA-CUSUM consistently achieves faster detection with smaller false alarm rates
than non-adaptive baselines, while outperforming existing UQ- and vision-based OOD methods at lower computational cost.

\item We instantiate MA-CUSUM under full, partial, and
unknown knowledge of the post-change distribution, making the
detector deployment-ready when the distribution is uncertain.

\item Our design is modular and compatible with diverse prediction models and error distributions, requiring only prediction residuals as input.
Computationally, our method only introduces a one-timestamp delay beyond the inherent detection delay, making it practical for real-time use.
\end{itemize}

\section{Related Work}
\noindent
{\bf Trajectory Prediction.}
\label{subsec:trajectory_prediction}
Trajectory prediction is a critical component in autonomous driving.
Early approaches relied on recurrent neural networks~\cite{zyner2018recurrent}
to model single-agent behavior,
while subsequent work introduced interaction-aware mechanisms such as social pooling and graph-based models to capture multi-agent dependencies~\cite{li2019grip}.
More recently, advanced methods have increasingly adopted Transformer architectures~\cite{postnikov2021transformer} and spatio-temporal graph neural networks~\cite{wang2023data} to improve accuracy, enable real-time forecasting, and capture complex interactions.
Despite these advances, the reliability of trajectory prediction under distribution shifts remains underexplored.
A small number of studies address related issues through uncertainty estimation~\cite{wiederer2023joint} and adversarial attacks~\cite{zhang2022adversarial}.
Robustness to OOD is increasingly recognized as a fundamental performance indicator for trajectory prediction models \cite{wiederer2023joint}.
However, comprehensive approaches for OOD detection in trajectory prediction remain underexplored. Our work specifically investigates OOD awareness in trajectory prediction.

\noindent{\bf OOD Detection in AVs.}
Existing OOD detection research in AVs has primarily focused on perception tasks such as object detection, semantic segmentation, and depth estimation~\cite{9564545, shoeb2025out}.
Representative approaches include scoring functions such as maximum softmax probability (MSP)~\cite{hendrycks2016baseline} and energy-based scores~\cite{liu2020energy}, as well as latent-space density estimation techniques~\cite{lee2018simple}.
These OOD detection methods are not fully suited for trajectory prediction since they
overlook the spatiotemporal correlations among the trajectories of surrounding agents.
Moreover, as summarized in~\cite{cui2022out}, prevailing OOD performance metrics often prioritize recognition accuracy without addressing detection delay and computation complexity.
In AV practice, short detection delay is critical for enabling timely control adaptation, while low computational complexity is necessary to satisfy on-board resource constraints.

\noindent{\bf Application of UQ to OOD Detection.} Popular UQ methods for trajectory prediction~\cite{gal2016dropout, nayak2022uncertainty} include deep ensembles, Monte Carlo dropout, and Bayesian neural networks. While effective, these methods typically incur substantial computational overhead.
Recent research has focused on developing more efficient UQ frameworks tailored for OOD detection. For instance,
\cite{wiederer2023joint} introduced a scalable approach that models ID samples within a latent space using a Gaussian Mixture Model (GMM), identifying OOD samples as low-likelihood outliers---a principle shared by other density-based detection methods~\cite{lee2018simple,sun2022out}.
Complementarily, \cite{10.1145/3803009} introduced an external monitoring module that operates directly on model outputs, detecting OOD samples via prediction error statistics without altering the underlying predictor.

\noindent{\bf Quickest Change-Point Detection (QCD). }
\label{subsec:qcd}
QCD provides a rigorous framework for monitoring sequential data to identify the  \textit{change-point} at which its underlying distribution shifts, with the dual objectives of minimizing the detection delay and controlling the frequency of false alarms \cite{veeravalli2014quickest}. This formalism, with applications spanning finance, climatology, and network monitoring \cite{tartakovsky2012efficient}, is typically pursued within either a Bayesian~\cite{tartakovsky2014sequential} or minimax setting. In particular, the CUSUM algorithm is first-order asymptotically optimal \cite{moustakides1986optimal, lorden1971procedures, veeravalli2014quickest}.
These strong theoretical guarantees, combined with its computational
efficiency, make CUSUM a powerful and principled foundation for our work.

\section{Mode-Dependent Error Distributions}
\label{sec: multiple modes}

\begin{figure*}[t!]
\vspace{0.4em}
    \centering
    \includegraphics[width=0.9\linewidth]{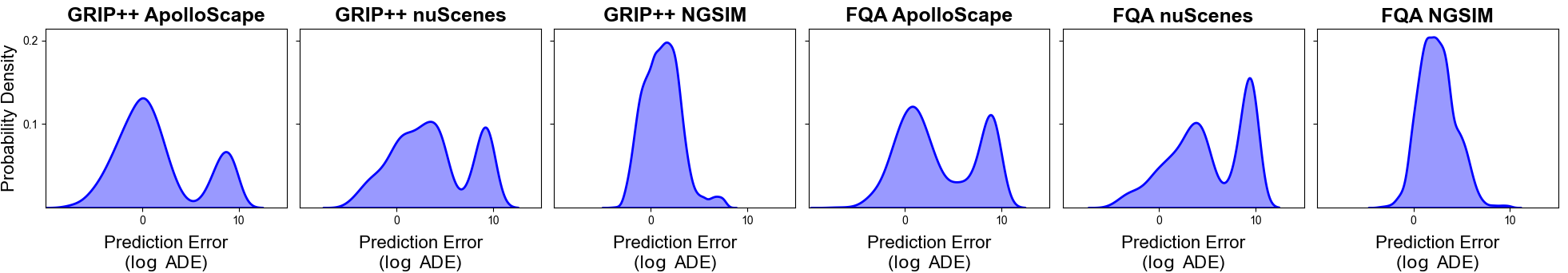}
    \caption{\textbf{Multi-modal distributions of prediction errors.}
    The $x$-axis denotes prediction error (log-scale ADE; Sec.~\ref{sec:metrics}), and the $y$-axis denotes probability density.
    Each plot is a kernel density estimate pooled over $500$ independently sampled test sequences; across all draws, the empirical distribution consistently exhibits a bi-modal structure, well approximated by a two-component GMM ($K\!=\!2$, fitted on ID data).
    The lower-error component is defined as the \emph{low-risk mode}, while the higher-error component represents the \emph{high-risk mode}. }
    \label{fig:GMM}
    \vspace{-1em}
\end{figure*}

Our empirical analysis across multiple representative real-world benchmarks
shows that prediction errors consistently exhibit \textit{multi-modal} distributions whose {active mode evolves over time}.
We conjecture that this multi-modality behavior is not dataset-specific but generalizes to a broader spectrum of trajectory prediction models and datasets. Moreover, the set of modes may extend beyond two, potentially encompassing a richer class of error regimes.

\begin{figure}[t]
    \centering
    \begin{minipage}{0.16\textwidth}
        \includegraphics[width=\linewidth]{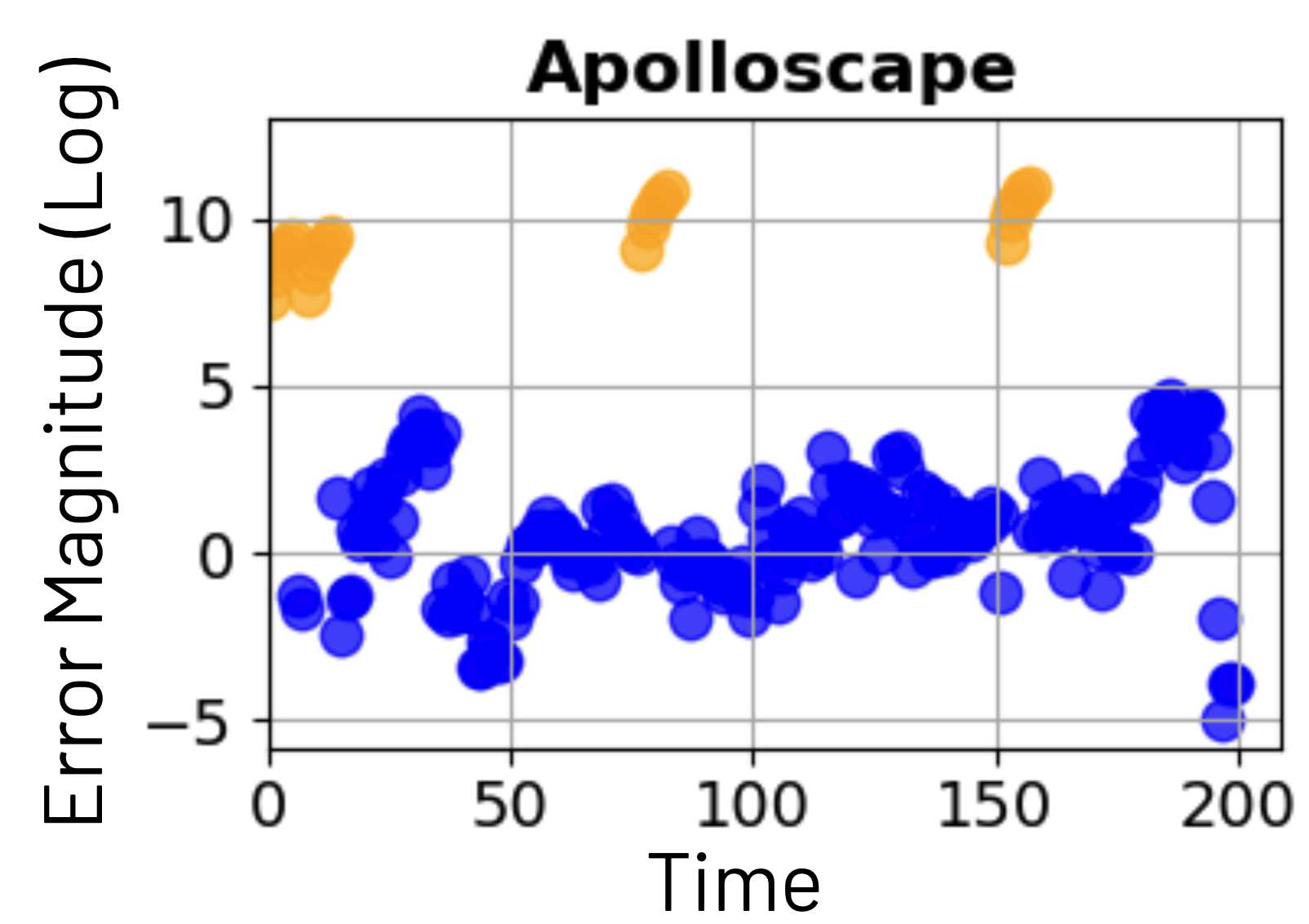}
    \end{minipage}
    \hspace{-0.5em}
    \begin{minipage}{0.16\textwidth}
        \includegraphics[width=\linewidth]{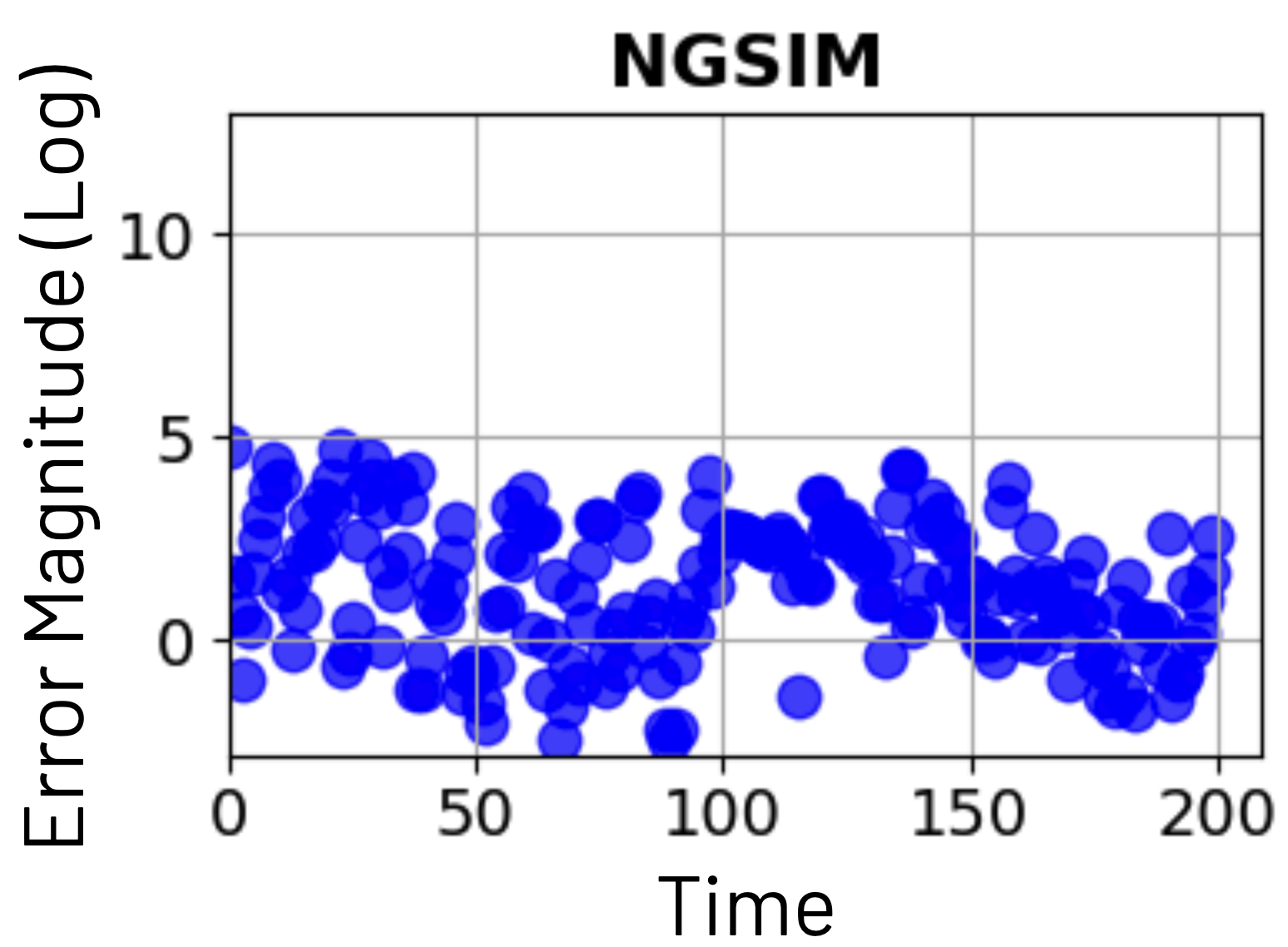}
    \end{minipage}
    \hspace{-0.5em}
    \begin{minipage}{0.16\textwidth}
        \includegraphics[width=\linewidth]{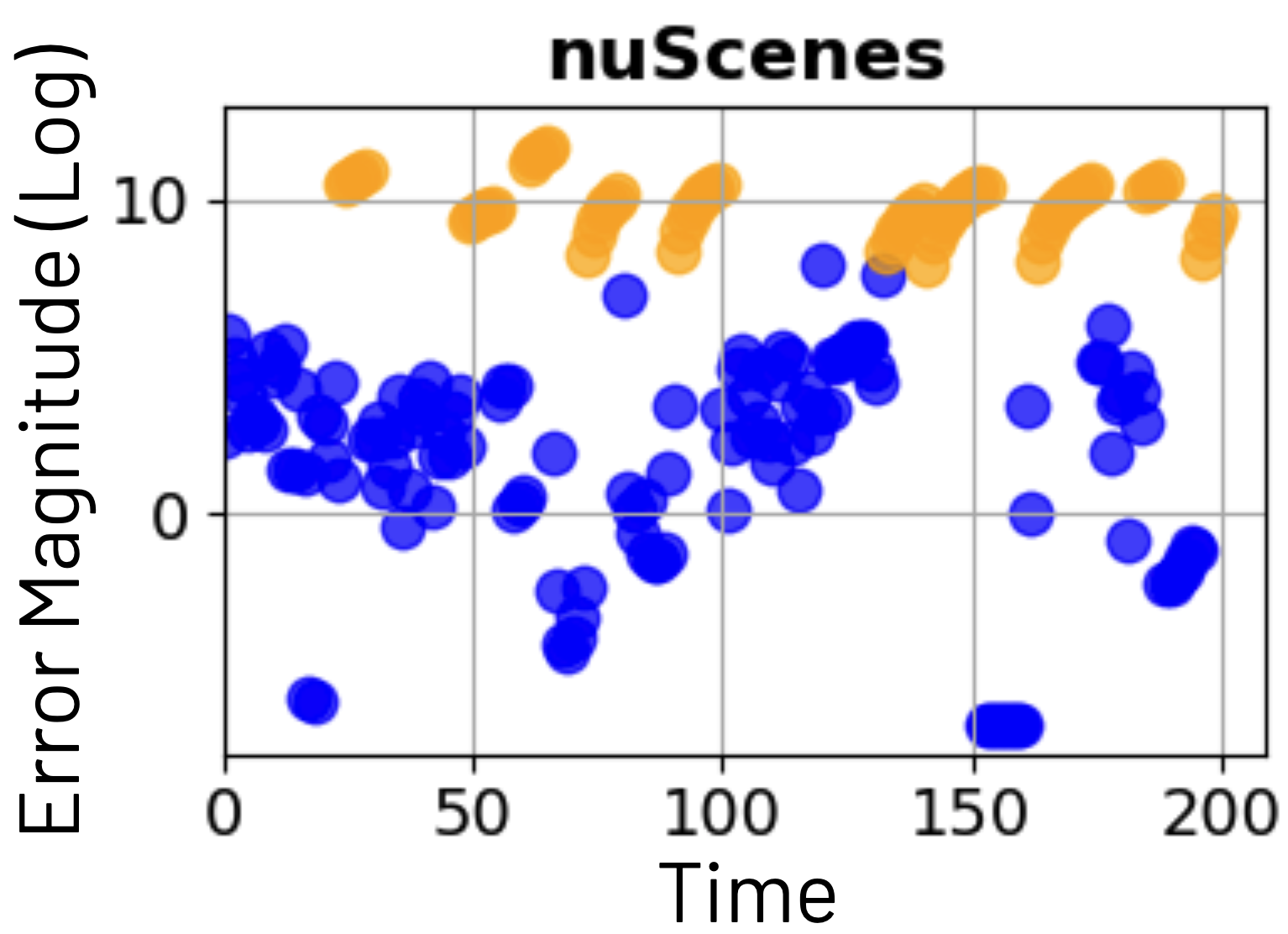}
    \end{minipage}
    \caption{\textbf{Mode dynamics across different driving scenes (FQA model).} Datasets: \textit{ApolloScape} (left), \textit{NGSIM} (center), \textit{nuScenes} (right). The $x$-axis represents the time frame indices; the $y$-axis is error magnitude (log scale).
    The modes were inferred by \cite{hajek2015random} via the GMM posterior (\textcolor{blue}{\texttt{blue}} = low-risk, \textcolor{orange}{\texttt{orange}} = high-risk).
    }
    \label{fig:mode_dynamics}
    \vspace{-15pt}
\end{figure}

\begin{table}[htbp]
\centering
\caption{{GMM mode weights (\%).}}
\label{tab:mode_weights}
\footnotesize
\setlength{\heavyrulewidth}{0.08em}
\setlength{\lightrulewidth}{0.04em}
\begin{tabular}{@{}l c c c@{}}
\toprule
& \textbf{ApolloScape} & \textbf{nuScenes} & \textbf{NGSIM} \\
\midrule
GRIP++ & 77.5\,/\,22.5 & 67.4\,/\,32.6 & 94.1\,/\,5.9 \\
FQA    & 57.7\,/\,42.3 & 60.2\,/\,39.8  & 80.8\,/\,19.2 \\
\bottomrule
\multicolumn{4}{@{}l}{\scriptsize Low-risk\,/\,high-risk.}
\end{tabular}
\vspace{-2em}
\end{table}

\begin{figure*}[t]
\vspace{0.4em}
    \centering
    \begin{minipage}{0.32\linewidth}
        \includegraphics[width=\linewidth]{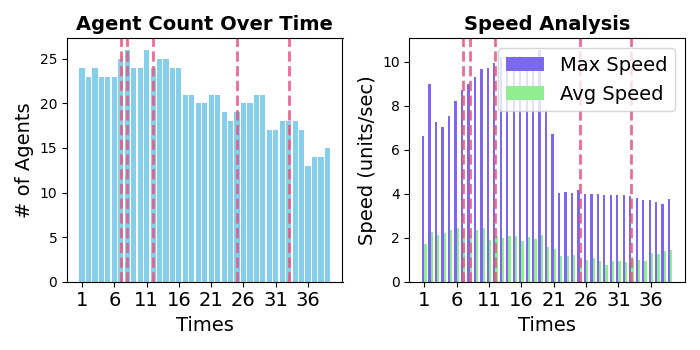}
    \end{minipage}
    \begin{minipage}{0.65\linewidth}
        \includegraphics[width=\linewidth]{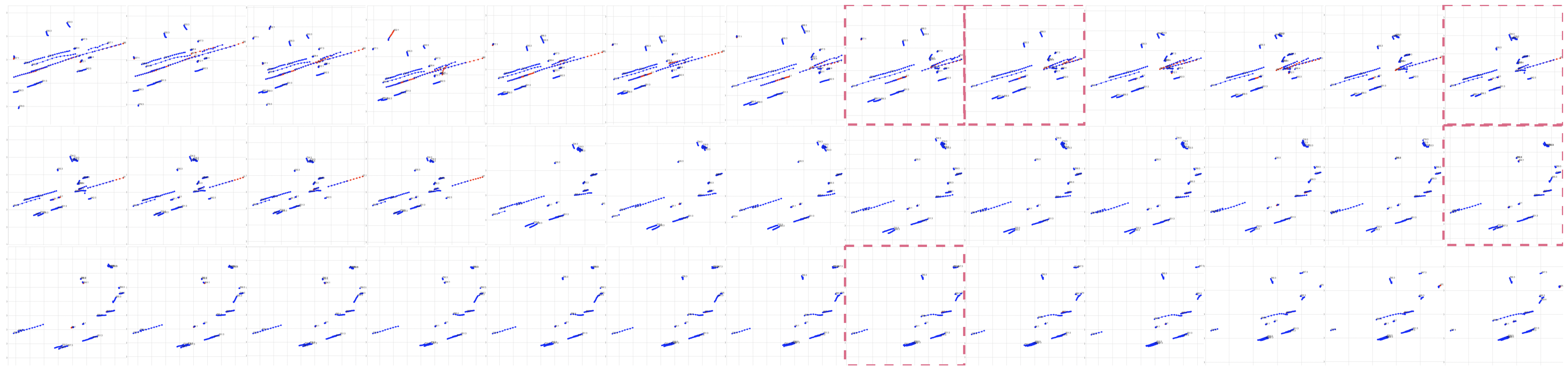}
    \end{minipage}
    \caption{\textbf{Scene-level observables provide unreliable cues for detecting latent-mode transitions.}
     We plot agent count and maximum/average speed against time, alongside map snapshots of the corresponding scenes --- each small tile shows the local map at the current frame, capturing the transition from a straight-lane segment into a left turn. \textcolor[rgb]{0.858,0.188,0.478}{\texttt{Pink}} dashed outlines mark the frames at which a mode switch occurs. At every transition, agent count and speed profiles change negligibly and the scene appearance is visually indistinguishable from neighboring frames. This gap between what is observable and what is detectable motivates a statistics-based mode estimator rather than one driven by scene appearance or hand-crafted observables.
    }
    \label{fig:mode_switch}
    \vspace{-1em}
\end{figure*}

\subsection{Overall Multiple Error Modes}
\label{subsec: multiple error modes}
We begin with an empirical study of prediction errors across multiple models (e.g., GRIP++~\cite{li2019grip++}, FQA~\cite{kamra2020multi}) and datasets (ApolloScape~\cite{huang2018ApolloScape}, NGSIM~\cite{fhwa2020}, nuScenes~\cite{caesar2020nuscenes}).
The datasets vary in map topology, traffic composition, and density; details can be found in Section \ref{subsec: experiment setup}.

As illustrated in Fig.~\ref{fig:GMM}, the overall prediction errors for each of the datasets are well approximated by a Gaussian Mixture Model (GMM) that naturally divides into two latent components: a \emph{low-risk mode} component and a \emph{high-risk mode} component. Table~\ref{tab:mode_weights} reports the GMM mode weights across all model--dataset combinations. The high-risk mode proportion varies substantially---from 5.9\% to 42.3\%---reflecting the diversity of driving contexts and models.

\vspace{-0.5em}
\subsection{On the Temporal Evolution of Error Modes}
\label{subsec: mode evolution}
We observe that the error modes evolve over time, potentially following different dynamics across datasets.
Intuitively, in highway driving, the ego vehicle often remains in the same error mode for extended periods with occasional transitions, whereas dense urban intersections feature more abrupt and frequent behavioral changes, resulting in frequent mode switches.
Our experimental results align with this intuition, as is shown in Fig. \ref{fig:mode_dynamics}:
\begin{itemize}
    \item \textbf{ApolloScape and nuScenes:} Urban intersection settings in which close-proximity interactions drive frequent and abrupt mode shifts.
    \item \textbf{NGSIM:} Highway settings in which prediction errors predominantly remain in the low-risk mode, with occasional high-risk episodes corresponding to maneuvers such as lane changes or merging.
\end{itemize}
The mode dynamics under GRIP++ are expected to be somewhat similar to those in Fig.\,\ref{fig:mode_dynamics} yet with overall lower frequency in the high-risk mode in ApolloScape and nuScenes.

\subsection{Latent Modes are Not Directly Observable}
Due to the inherent randomness of prediction errors, the underlying error mode is often latent and not directly observable from visual inputs.
Readily available scene-level statistics---such as agent counts, maximum/average speeds, or map snapshots---offer no reliable cues for identifying mode transitions.
As shown in Fig.~\ref{fig:mode_switch}, mode transitions (highlighted by dashed pink lines) occur without significant corresponding changes in observable metrics: agent density remains stable, speed metrics show no anomalous patterns, and scene snapshots display consistent visual characteristics.
The weak correspondence between observable scene information and error modes limits the effectiveness of purely vision-based approaches for mode identification, thereby motivating our statistics-based mode estimation framework.

\vspace{-0.2em}
\section{Problem Formulation: Multi-Modal QCD}
\label{sec: pilot}
Observing the near-Gaussian mixture distributions with two modes in Fig. \ref{fig:GMM}, we formulate the problem for the two-mode case. Our formulation can be naturally extended to an arbitrary number of modes.
\vspace{-0.5em}
\subsection{Overall Error Distribution}
Let $\epsilon$ denote the prediction error.
Let $f(\cdot)$ denote the overall error distribution of the training dataset, i.e., the empirical distribution function.  For ease of exposition, we assume the training dataset is sufficiently large so that $f$ can be treated as a probability density function (pdf).
Let $\mathcal{M} = \{0, 1\}$ denote the two possible modes, with $0$ indicating the low-error mode and $1$ indicating the high-error mode.
Suppose that $f$ is a mixture distribution of two components, formally
$f (\epsilon)= \pi_0 f_0 (\epsilon) + \pi_1 f_1 (\epsilon)$,
where $\pi_0 \ge 0$, $\pi_1 \ge 0$, and $\pi_0+\pi_1 = 1$.
The supports of $f_0$ and $f_1$ may overlap with each other. For example, when $f$ is a Gaussian mixture, the supports of $f_0$ and $f_1$ are both the entire $\mathbb{R}$.
Without loss of generality, we assume $\mathbb{E}_{\epsilon\sim f_0}[\epsilon] < \mathbb{E}_{\epsilon\sim f_1}[\epsilon]$.

\subsection{Mode Dynamics}
\label{sec:multiple_modes}

Let $M_t \in \mathcal{M}$ denote the latent error mode at time $t$. The mode is \emph{unobservable} and its \emph{temporal evolution is unknown a priori}.
In the real world, under the same overall error distribution $f$, there can be a wide range of mode dynamics.
Some representative examples are as follows.

\vspace{0.1em}
\noindent{\em (i) Static.} The mode $M_t$ remains in a constant state $m^\star$ for extended durations, satisfying $P(M_t = M_{t-1}) \approx 1$. With initial distribution $\mathbb{P}(M_0 = 0) = \pi_0$, this dynamic captures highly persistent behaviors---such as highway cruising in \emph{NGSIM}---where state switches are infrequent.

\vspace{0.1em}
\noindent {\em (ii) i.i.d. switching.} $\mathbb{P}(M_t=m)=\pi_m$ for $m\in \{0,1\}$ and the $M_t$'s are independent across time.
This dynamic encodes memoryless fluctuations.

\vspace{0.1em}
\noindent {\em (iii) Markovian,} i.e., $\mathbb{P}\{M_t\mid M_{t-1}, \ldots, M_0\} = \mathbb{P}\{M_t\mid M_{t-1}\}$ for any $t\ge 1$.  This dynamic formalizes temporally correlated transitions, which typically occur in structured urban scenes.
\vspace{0.1em}
\noindent {\em (iv) Arbitrary deterministic,} where the evolution of $M_t$ follows no well-studied random process. This is the most general and pessimistic setting, though we conjecture that real-world traffic modes typically retain exploitable temporal structure.

Unifying these dynamics through the overall distribution $f$, our framework captures the diversity and evolution of real-world traffic with analytical rigor.

\subsection{OOD Detection as Change-point Detection}

Let $\{\epsilon_t\}_{t=1}^{\infty}$ denote the sequence of prediction errors. Let $\gamma$ be the unknown time at which the ego vehicle enters an OOD scenario, triggering a distributional shift:
\[
    \epsilon_t \sim
    \begin{cases}
        f_t(\cdot), & t < \gamma \\
        g(\cdot), & t \geq \gamma.
    \end{cases}
\]
Here, $f_t$ represents the ID density.
When $M_t = 0$, $f_t = f_0$, and $f_t = f_1$ otherwise.
The post-change distribution $g(\cdot)$ represents the OOD density.

\subsubsection{\underline{Detection Objective}}
For any detection algorithm, we denote by $\tau$ the time it declares OOD.
For robustness, we adopt Lorden's minimax criterion~\cite{lorden1971procedures}, and define the \emph{Worst-Case Average Detection Delay (WADD)} as
\[
    \text{WADD}(\tau) = \sup_{t \geq 1} \, \text{ess} \sup \; \mathbb{E}_t\!\left[ \, (\tau - t + 1)^+ \mid \epsilon_1,\dots,\epsilon_{t-1} \, \right],
\]
where $(x)^+ = \max\{0,x\}$ and $\mathbb{E}_t[\cdot]$ denotes the expectation when the change occurs at $t$.

False alarms are quantified by the \emph{False Alarm Rate (FAR)} and its reciprocal, the \emph{Mean Time to False Alarm (MTFA)}:
\[
    \text{FAR}(\tau) = \frac{1}{\mathbb{E}_\infty[\tau]},
    \qquad
    \text{MTFA}(\tau) = \mathbb{E}_\infty[\tau],
\]
where $\mathbb{E}_\infty$ denotes expectation under the null hypothesis (no change occurs).

\subsubsection{\underline{Classic CUSUM Procedure}}
For the simplified setting where $f_t = \tilde{f}$ is time-invariant, and both $\tilde{f}$ and $g$ are known, the classic CUSUM procedure \cite{page1954continuous} is optimal for minimizing WADD \cite{moustakides1986optimal}. Intuitively, the test statistic $S_t$ accumulates the log-likelihood ratio via
\begin{align}
\label{eq:cusum}
    S_t = \max\left(0,\; S_{t-1} + \ln \frac{g(\epsilon_t)}{\tilde{f}(\epsilon_t)}\right), \quad S_0 = 0.
\end{align}
A change is declared at $\tau = \inf\{t \geq 1 : S_t \geq b\}$, where the threshold $b$ is calibrated to satisfy the given MTFA constraint.

\subsubsection{\underline{Operational Knowledge Paradigms}}
\label{sec:knowledge_settings}
In real-world practice, the a priori knowledge of $f_t$ and $g$ can vary.
Oftentimes, the overall error distribution $f$ can be estimated from ID data. Hence, we assume $f$ is known. We consider multiple knowledge settings for the post-change distributions, following~\cite{10.1145/3803009}.

\noindent {\bf Setting I: Full Knowledge.}
The post-change distribution is fully specified.
This setting serves as a performance upper bound.
In many applications, $g$ may be approximated via domain knowledge, incident logs, or simulation.

\noindent\textbf{Setting II: Partial Knowledge.}
When only coarse statistics of $g$ (e.g., mean $\mu_g$, variance $\sigma_g^2$) are available, we approximate $g$ by $\hat{g}(\epsilon) = \mathcal{N}(\epsilon \mid \mu_g, \sigma_g^2)$ and substitute into~Eq.\,\eqref{eq:cusum}. This approximation remains valid as long as the log-likelihood ratio has negative drift before the true change-point and positive drift after it.

\noindent\textbf{Setting III: Unknown Knowledge.}
When no parametric form for $g$ is available, we adopt a robust shift-based CUSUM procedure. The post-change density is approximated by the location-shifted pre-change density $f(\epsilon - \kappa)$, where $\kappa > 0$ is a user-specified \emph{minimum detectable shift}, yielding the statistic
\begin{equation}
\label{eq:shift_cusum}
  S_t^{(\kappa)} = \max\left(0, \; S_{t-1}^{(\kappa)} + \ln\frac{f(\epsilon_t - \kappa)}{f(\epsilon_t)}\right).
\end{equation}
Since the error metrics (Sec.~\ref{sec:metrics}) are non-negative, $f$ is supported on $\mathbb{R}_{\geq 0}$. The shifted family $\{f(\cdot - \eta) : \eta \ge \kappa\}$ satisfies the weak stochastic boundedness conditions of \cite{molloy2017misspecified}. Thus, by Theorem~3 therein, the test is asymptotically minimax robust over this family, guaranteeing valid false-alarm control for any true shift $\eta \ge \kappa$ \cite{10.1145/3803009}.
Following the guidelines in~\cite{10.1145/3803009} and for fair comparison, we set $\kappa = 1$ in all subsequent experiments.

\subsection{Prediction Error Metrics}
\label{sec:metrics}
In trajectory prediction, $\epsilon_t$ may be defined in different ways depending on the chosen evaluation metric. We adopt three standard metrics~\cite{kim2020multi}:
\begin{itemize}
    \item \emph{Average Displacement Error (ADE):} mean $\ell_2$ distance between predicted and ground-truth positions across the prediction horizon.
    \item \emph{Final Displacement Error (FDE):} $\ell_2$ distance at the final prediction step, highlighting long-horizon accuracy.
    \item \emph{Root Mean Squared Error (RMSE):} square root of the average squared deviation, which penalizes large errors and is sensitive to outliers or abrupt shifts.
\end{itemize}

\section{Method}
\label{sec:method}

Algorithm~\ref{alg:mode_aware_cusum} processes the prediction error
stream $\epsilon_1, \epsilon_2, \ldots$ sequentially. At each step $t$,
three operations are performed:

\noindent
\ding{172} {\it Mode Estimation:}
Because the two GMM components share overlapping support
(Fig.~\ref{fig:GMM}), a hard partition of the error space is
unavailable. We therefore infer the latent mode $M_t \in \{0,1\}$ via
Maximum A Posteriori (MAP) estimation~\cite{hajek2015random}:
$
\hat{M}_t = \arg\max_{m}\, P(M_t = m \mid \epsilon_t),
P(M_t = m \mid \epsilon_t) = \frac{\pi_m\, f_m(\epsilon_t)}{f(\epsilon_t)},
$
where $\pi_m$ is the prior weight of mode $m$ and $f_m$ its mixture
component. Since the denominator $f(\epsilon_t)$ is common to both modes,
the MAP rule reduces to $\hat{M}_t = \arg\max_{m}\, \pi_m f_m(\epsilon_t)$.

This estimator has three deployment-relevant properties:
\begin{itemize}
    \item \textbf{Efficient:} uses only \(\epsilon_t\), costing \(O(1)\) per step;
    \item \textbf{Stateless:} requires neither future observations nor a warm-up window;
    \item \textbf{Bayes-optimal:} under 0-1 loss~\cite{cover1999elements}, so no mode-transition model need be assumed.
\end{itemize}

\begin{remark}[{\it Robustness to isolated misclassifications}]
A single misclassification perturbs one LLR increment, which subsequent accumulation absorbs; $r_0 > r_1$ further keeps the low-error threshold conservative.
\end{remark}

\noindent
\ding{173} {\it Adaptive Threshold Update:}
The threshold combines a local variance estimate with a bound derived from Wald's Sequential Probability Ratio Test (SPRT)~\cite{wald1947sequential}. The construction is decomposed into four steps.

\begin{itemize}
    \item \textbf{Local variance.}
    Let \(\mathcal{I}_t = \{\tau \leq t : \hat{M}_\tau = m\}\) index past steps assigned to mode \(m\).
    Define \(\hat{\sigma}^{(m)}\) as the sample standard deviation of the residuals \(\{\epsilon_\tau\}\) over the \(L\) most recent indices in \(\mathcal{I}_t\).

    \item \textbf{Mode-specific signal scale.}
    For a given mode \(m\), the smallest shift considered meaningful relative to within-mode noise is
    \(d_m = r_m \, \hat{\sigma}^{(m)}\), \(r_m > 0\),
    where the multiplier \(r_m\) is a per-mode tuning knob.

    \item \textbf{Base threshold via SPRT.}
    Wald's SPRT gives the upper likelihood-ratio boundary \(A = (1-\beta)/\alpha\).
    Scaling its log-domain form by the per-step Kullback--Leibler divergence \(d_m^{2}/2\)~\cite{lorden1971procedures, veeravalli2014quickest} yields the threshold
    \(h_m = \frac{2}{d_m^{2}} \ln\!\left(\frac{1-\beta_m}{\alpha_m}\right)\),
    where \(\alpha_m\) and \(\beta_m\) are the per-mode target false-alarm and missed-detection rates.

    \item \textbf{Smoothing.}
    The instantaneous threshold \(h_m\) fluctuates with \(\hat{\sigma}^{(m)}\) as new errors arrive.
    To suppress this jitter while preserving responsiveness to genuine regime change, we apply an Exponentially Weighted Moving Average (EWMA)~\cite{roberts1959control}:
    \(\theta_t^{(m)} = \lambda\, h_m + (1-\lambda)\,\theta_{t-1}^{(m)}\), \(\lambda \in (0,1]\).
Smoothing factor $\lambda$ governs the tracking--stability trade-off: larger $\lambda$ prioritizes the current $h_m$ for faster adaptation; smaller $\lambda$ favors history for greater stability. EWMA's geometric weighting ensures $O(1)$ memory and update cost, ideal for the strictly online setting.
\end{itemize}

\begin{remark}[{\it Mode-dependent threshold calibration}]
Since \(\hat{\sigma}^{(0)} \ll \hat{\sigma}^{(1)}\) in our setting, selecting \(r_0 > r_1\) moderates \(h_0\) to an operationally useful level, preventing the threshold from becoming excessively restrictive for the dominant mode.
\end{remark}

\setlength{\textfloatsep}{4pt plus 2pt minus 4pt}
\begin{algorithm}[t]
\footnotesize
\KwIn{%
  Pre-change GMM $(\pi_0,\pi_1,f_0,f_1)$\;
  Knowledge $\mathcal{K}\!\in\!\{\textsc{Full},\textsc{Partial},\textsc{Unknown}\}$
  with post-change specifications:
  \quad $g(\cdot)$ if $\mathcal{K}=$Full;
  $(\mu_g,\sigma_g^2)$ if $\mathcal{K}=$Partial. \\
  Tuning: initial thresholds $\theta_0^{(m)}$, multipliers $r_m$, target rates $\alpha_m,\beta_m$ for $m\in\{0,1\}$,
  window~$L$, smoothing~$\lambda$\;
}
\KwOut{Change point $\tau$}
\BlankLine
Init.\ $S_0^{(m)}\!\gets\!0$\;
$\theta_0^{(m)}$ given for $m\!\in\!\{0,1\}$\;
\For{$t = 1, 2, \ldots$}{
  Compute prediction error $\epsilon_t$\;
  \tcc{Step 1: Mode estimation (MAP)}
  $\hat{M}_t \gets \arg\max_{m\in \{0,1\}}\,\pi_m f_m(\epsilon_t)$\;
  \tcc{Step 2: Adaptive threshold}
  Estimate $\hat{\sigma}^{(\hat{M}_t)}$ over window~$L$\;
  $d_{\hat{M}_t} \gets r_{\hat{M}_t}\hat{\sigma}^{(\hat{M}_t)}$\;
  $h_{\hat{M}_t} \gets \frac{2}{d_{\hat{M}_t}^2}
    \ln\!\bigl(\frac{1-\beta_{\hat{M}_t}}{\alpha_{\hat{M}_t}}\bigr)$\;
  $\theta_t^{(\hat{M}_t)}
    \gets \lambda h_{\hat{M}_t}
    +(1-\lambda)\theta_{t-1}^{(\hat{M}_t)}$\;
  \tcc{Step 3: Knowledge-aware LLR}
  \uIf(\tcp*[f]{Setting I}){$\mathcal{K}=\textsc{Full}$}{
    $\ell_t \gets \ln\frac{g(\epsilon_t)}{f_{\hat{M}_t}(\epsilon_t)}$}
  \uElseIf(\tcp*[f]{Setting II}){$\mathcal{K}=\textsc{Partial}$}{
    $\ell_t \gets \ln\frac{\mathcal{N}(\epsilon_t\mid\mu_g,\sigma_g^2)}{f_{\hat{M}_t}(\epsilon_t)}$}
  \Else(\tcp*[f]{Setting III}){
    $\ell_t \gets \ln\frac{f_{\hat{M}_t}(\epsilon_t-\kappa)}{f_{\hat{M}_t}(\epsilon_t)}$}
  \tcc{Step 4: CUSUM update}
  $S_t^{(\hat{M}_t)}
    \gets \max\!\bigl(S_{t-1}^{(\hat{M}_t)}
    +\ell_t,\;0\bigr)$\;
  \lIf{$S_t^{(\hat{M}_t)} \ge \theta_t^{(\hat{M}_t)}$}{\Return $\tau=t$}
}
\caption{Mode-Aware CUSUM (MA-CUSUM)}
\label{alg:mode_aware_cusum}
\end{algorithm}

\noindent
\ding{174} {\it CUSUM Update and Detection:}
At each step, the log-likelihood ratio $\ell_t$ quantifies shift evidence according to the knowledge paradigm: \textbf{Setting~I:}
    $\ell_t = \ln\bigl(g(\epsilon_t)\,/\,f_{\hat{M}_t}(\epsilon_t)\bigr)$;
\textbf{Setting~II:}
    $\ell_t = \ln\bigl(\mathcal{N}(\epsilon_t\mid\mu_g,\sigma_g^2)\,/\,
    f_{\hat{M}_t}(\epsilon_t)\bigr)$, using a Gaussian surrogate;
\textbf{Setting~III:}
    $\ell_t = \ln\bigl(f_{\hat{M}_t}(\epsilon_t-\kappa)\,/\,
    f_{\hat{M}_t}(\epsilon_t)\bigr)$, with shift
    $\kappa=1$~\cite{10.1145/3803009}.
The CUSUM statistic accumulates this evidence via
\begin{equation}
\label{eq:ma_cusum_update}
  S_t^{(\hat{M}_t)} \;=\; \max\!\bigl(
    S_{t-1}^{(\hat{M}_t)} + \ell_t,\; 0 \bigr),
\end{equation}
which retains the standard recursion of the classical
CUSUM~\cite{page1954continuous}: the log-likelihood ratio $\ell_t$ has
negative expectation before the change-point and positive expectation
after it, so no explicit drift subtraction is required. The
mode-specific signal scale $d_{\hat{M}_t}=r_{\hat{M}_t}\hat{\sigma}^{(\hat{M}_t)}$
enters the detector only through the adaptive threshold $\theta_t^{(\hat{M}_t)}$
(via $h_{\hat{M}_t}$), which is where all mode-dependent calibration
resides. A change is declared
when $S_t^{(\hat{M}_t)} \ge \theta_t^{(\hat{M}_t)}$.

\begin{remark}[{\it Post-detection restart}]
On OOD detection at \(\tau\), both statistics reset to zero and monitoring resumes from \(\tau+1\). This standard CUSUM restart handles multiple OOD episodes within a single drive. Statistics update independently per mode, but only the current mode \(\hat{M}_t\) receives the LLR increment \(\ell_t\). Formal restart and false-alarm guarantees under mode-switching appear in our companion work~\cite{guo2026latent}.
\end{remark}

\section{Experiments}
\label{sec:experiments}
\subsection{Experimental Setup}
\label{subsec: experiment setup}

\begin{table*}[t]
\vspace{0.4em}
\centering
\caption{{OOD detection results (mean $\pm$ std.) on GRIP++ and FQA.} Our \textbf{MA-CUSUM} achieves the best \textbf{AUROC} {(ranking accuracy)} and \textbf{AUPR} {(precision--recall robustness under class imbalance)}, consistently surpassing all baselines. {Compared to Global CUSUM~\cite{10.1145/3803009}, it also produces markedly fewer false positives at high recall.}}
\label{tab:results}
\vspace{0.3em}
\begin{tabular}{ll|cc|cc}
\toprule
& \multirow{2}{*}{Method} & \multicolumn{2}{c|}{\textbf{GRIP++}} & \multicolumn{2}{c}{\textbf{FQA}} \\
& & AUROC & AUPR($10^{-1}$) & AUROC & AUPR($10^{-1}$) \\
\midrule
\multirow{2}{*}{\textit{Likelihood-based}}
& lGMM~\cite{wiederer2023joint} & $0.78 \pm 0.01$ & $0.31 \pm 0.02$ & $0.81 \pm 0.00$ & $0.37 \pm 0.01$ \\
& NLL~\cite{lee2018simple} & $0.48 \pm 0.00$ & $0.08 \pm 0.00$ & $0.53 \pm 0.01$ & $0.18 \pm 0.00$ \\
\midrule
\multirow{4}{*}{\textit{Sequential}}
& Z-Score~\cite{shewhart1931economic} & $0.72 \pm 0.01$ & $0.29 \pm 0.02$ & $0.75 \pm 0.01$ & $0.34 \pm 0.02$ \\
& Chi-Square~\cite{hotelling1947multivariate} & $0.76 \pm 0.01$ & $0.34 \pm 0.02$ & $0.78 \pm 0.01$ & $0.40 \pm 0.02$ \\
& Global CUSUM~\cite{10.1145/3803009} & $0.87 \pm 0.01$ & $0.53 \pm 0.07$ & $0.89 \pm 0.01$ & $0.76 \pm 0.08$ \\
\rowcolor{babypurple}
& MA-CUSUM (ours) & $0.93 \pm 0.01$ & $1.11 \pm 0.14$ & $0.95 \pm 0.01$ & $1.95 \pm 0.17$ \\
\bottomrule
\end{tabular}
\vspace{-1em}
\end{table*}

We evaluate MA-CUSUM across three datasets~\cite{huang2018ApolloScape, fhwa2020, caesar2020nuscenes} (details in Sec. \ref{subsec: mode evolution}).
MA-CUSUM is architecturally agnostic to the underlying predictor; we therefore select two models for architectural diversity rather than recency: \textit{GRIP++}~\cite{li2019grip++}, a graph-based two-layer GRU strong on short- and long-horizon forecasts (using $D_{\mathrm{close}}=25$ ft as in the original work to filter distant agents), and \textit{FQA}~\cite{kamra2020multi}, which integrates fuzzy rules with learned attention for adaptive interaction weighting in complex urban scenes such as nuScenes.

MA-CUSUM has six tuning parameters: target rates $\alpha_m,\beta_m$ are application-driven ($\alpha_m=0.05$ for 5\% FAR); defaults $r_0=0.5$, $r_1=0.3$ (with $r_0>r_1$ to moderate $h_0$), $L=50$, and $\lambda=0.1$ are fixed across all experiments.

\begin{figure}[t]
    \centering
    \includegraphics[width=0.55\linewidth]{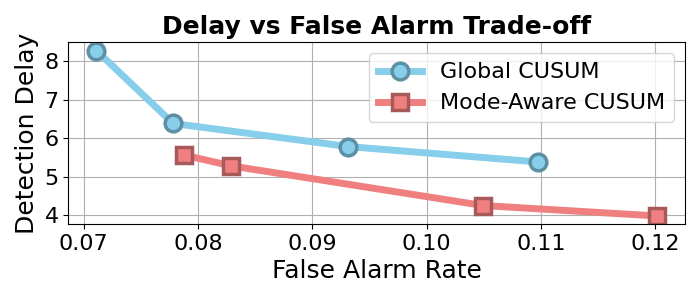}
    \caption{\textbf{Detection delay versus false alarm rate.} Lower values on both axes indicate better performance (faster detection, fewer false alarms). MA-CUSUM (\textcolor[rgb]{0.941,0.502,0.502}{\texttt{Red}}) shifts the trade-off curve toward the lower-left relative to \textit{Global CUSUM} (\textcolor[rgb]{0.529,0.808,0.922}{\texttt{Blue}}), achieving shorter detection delays at the same or lower false alarm rates.}
    \label{fig:delayvsfar}
\end{figure}

\subsection{OOD Data Generation}
\label{sec:ood_generation}

Evaluating runtime OOD detection requires realistic distribution shifts
with known change points. We adopt the physically-constrained adversarial
attack framework of~\cite{zhang2022adversarial} to generate OOD samples.
This approach perturbs observed trajectories while respecting vehicle
kinematics and scene constraints, yielding shifts that are subtle yet
capable of inducing large prediction errors---a realistic proxy for
safety-critical degradation. Unlike synthetic noise or manual trajectory
editing, these adversarial perturbations preserve scene semantics and
traffic interaction structure, producing challenging but physically
plausible distribution shifts.

\begin{table}[t!]
\centering
\scriptsize
\setlength{\tabcolsep}{2pt}
\caption{Trajectory Prediction Vulnerability: ID vs. OOD performance across ADE, FDE, and RMSE. All metrics exhibit significant degradation under adversarial shifts.}
\label{tab:ood_effectiveness}
\vspace{0.3em}
\begin{tabular}{ll|ccc|ccc|ccc}
\toprule
\multirow{2}{*}{\textbf{Model}} & \multirow{2}{*}{\textbf{Dataset}} & \multicolumn{3}{c|}{\textbf{ADE [m]}} & \multicolumn{3}{c|}{\textbf{FDE [m]}} & \multicolumn{3}{c}{\textbf{RMSE [m]}} \\
& & ID & OOD & $\Delta \%$ & ID & OOD & $\Delta \%$ & ID & OOD & $\Delta \%$ \\
\midrule
\multirow{3}{*}{GRIP} & Apollo & 1.97 & 7.14 & +262 & 3.42 & 12.48 & {+265} & 2.45 & 8.94  & +265 \\
                      & NGSIM  & 7.29 & 9.24 & +27  & 11.20 & 14.34 & +28  & 8.81 & 11.36 & +29  \\
                      & nuSc.  & 5.46 & 8.32 & +52  & 8.54  & 13.15 & +54  & 6.63 & 10.25 & +55  \\
\midrule
\multirow{3}{*}{FQA}  & Apollo & 2.37 & 5.64 & +138 & 4.15  & 9.88  & +138 & 3.02 & 7.18  & +138 \\
                      & NGSIM  & 5.44 & 7.05 & +30  & 9.62  & 12.60 & +31  & 7.15 & 9.38  & +31  \\
                      & nuSc.  & 5.28 & 7.83 & +48  & 8.25  & 12.21 & +48  & 6.55 & 9.72  & +48  \\
\bottomrule
\end{tabular}
\vspace{-1em}
\end{table}

\begin{table}[t!]
\vspace{0.4em}
\centering
\caption{
  Global (\textbf{G}) vs.\ MA-CUSUM (\textbf{M}) under four switching dynamics.
  WADD ($\downarrow$) and $\log_{10}(\text{MTFA})$ ($\uparrow$) are averaged over
  5{,}000 runs, with both detectors calibrated to a common false-alarm level.
  {\color{improvegreen}$\Delta$\%} denotes the relative improvement of M (ours) over G.
}
\label{tab:dynamics_performance}
\vspace{0.3em}
\begin{tabular}{l rrr@{\hspace{1.6em}} rrr}
\toprule
& \multicolumn{3}{c}{\textbf{WADD} $\downarrow$}
& \multicolumn{3}{c}{$\boldsymbol{\log_{10}}$\textbf{(MTFA)} $\uparrow$} \\
\cmidrule(lr){2-4}\cmidrule(lr){5-7}
\textbf{Dynamic}
  & \textbf{G} & \textbf{M} & {\color{improvegreen}$\boldsymbol{\Delta}$\textbf{\%}}
  & \textbf{G} & \textbf{M} & {\color{improvegreen}$\boldsymbol{\Delta}$\textbf{\%}} \\
\midrule
Static
  & 12.30 & 11.80 & {\color{improvegreen}$4.1$}
  & 3.45  & 3.48  & {\color{improvegreen}$7.2$}   \\
Markov
  & 18.70 & 11.20 & {\color{improvegreen}$40.1$}
  & 3.08  & 3.35  & {\color{improvegreen}$86.2$}  \\
Arbitrary
  & 17.20 & 12.00 & {\color{improvegreen}$30.2$}
  & 3.13  & 3.39  & {\color{improvegreen}$82.0$}  \\
i.i.d.
  & 13.80 & 12.90 & {\color{improvegreen}$14.6$}
  & 3.26  & 3.41  & {\color{improvegreen}$41.3$}  \\
\bottomrule
\end{tabular}
\end{table}

We apply perturbations starting at a predefined time,
giving us precise ground-truth change points for evaluating detection
delay (WADD) and false-alarm metrics without the ambiguity of
naturally occurring drift.
Table~\ref{tab:ood_effectiveness} summarises the impact of adversarial
perturbations on prediction accuracy across three datasets and
representative models, confirming that the generated OOD data
induces substantial and consistent performance degradation.

\subsection{Main Results}
\label{sec:results}

Unless stated otherwise, experiments use Setting I as a performance upper bound. Table~\ref{tab:robustness} demonstrates that while detection delay and AUROC degrade gracefully as knowledge diminishes, false-alarm control remains stable across all paradigms.

\vspace{0.7em}
\noindent\textbf{RQ1: Does {mode-awareness} improve OOD detection compared to likelihood-based and sequential baselines?}
{As shown in} Table~\ref{tab:results}, our method consistently outperforms {all} likelihood-based baselines (lGMM~\cite{wiederer2023joint}, NLL~\cite{lee2018simple}), sequential {detectors} (Z-Score~\cite{shewhart1931economic} and Chi-Square~\cite{hotelling1947multivariate}){,} and Global CUSUM~\cite{10.1145/3803009}. {On GRIP++, MA-CUSUM achieves 0.93 AUROC and 1.11 AUPR($\times 10^{-1}$), improving over Global CUSUM (0.87 / 0.53) by 6.9\% in AUROC and 109.4\% in AUPR. On FQA the gains are even larger: 0.95 vs.\ 0.89 AUROC (+6.7\%) and 1.95 vs.\ 0.76 AUPR ($\times 10^{-1}$, +156.6\%).}
{The large AUPR gains show that multi-mode scoring is especially effective at reducing false positives when OOD events are rare.}
{Overall, conditioning the sequential test on mode-specific error distributions consistently outperforms a single aggregate model in both ranking accuracy and precision.}

\vspace{0.7em}
\noindent{\bf RQ2: Does multi-mode improve the delay--false-alarm trade-off relative to single-distribution CUSUM?}
Beyond overall performance (RQ1), we examine whether {decomposing the detection statistic by mode} is necessary when prediction errors exhibit {mode-dependent heterogeneity}. Classical CUSUM assumes a single stationary pre-change distribution. {When traffic contains multiple behavioural modes (as stated in Section \ref{sec: multiple modes}), a single calibration induces a structural trade-off}: conservative thresholds delay detection of hazardous transitions, whereas aggressive thresholds inflate false alarms during benign intra-mode variability.
Fig.~\ref{fig:delayvsfar} shows that multi-mode consistently achieves lower detection delay at fixed false-alarm levels, with the largest gains in safety-relevant low-FAR regimes. The improvement arises because {mode-aware log-likelihood ratios preserve local statistical power}, whereas global averaging dilutes change evidence and enlarges the vulnerability window before alarm.

\begin{figure}[t]
    \centering
    \includegraphics[width=\linewidth, trim={2.6cm 0cm 4cm 2cm}, clip]{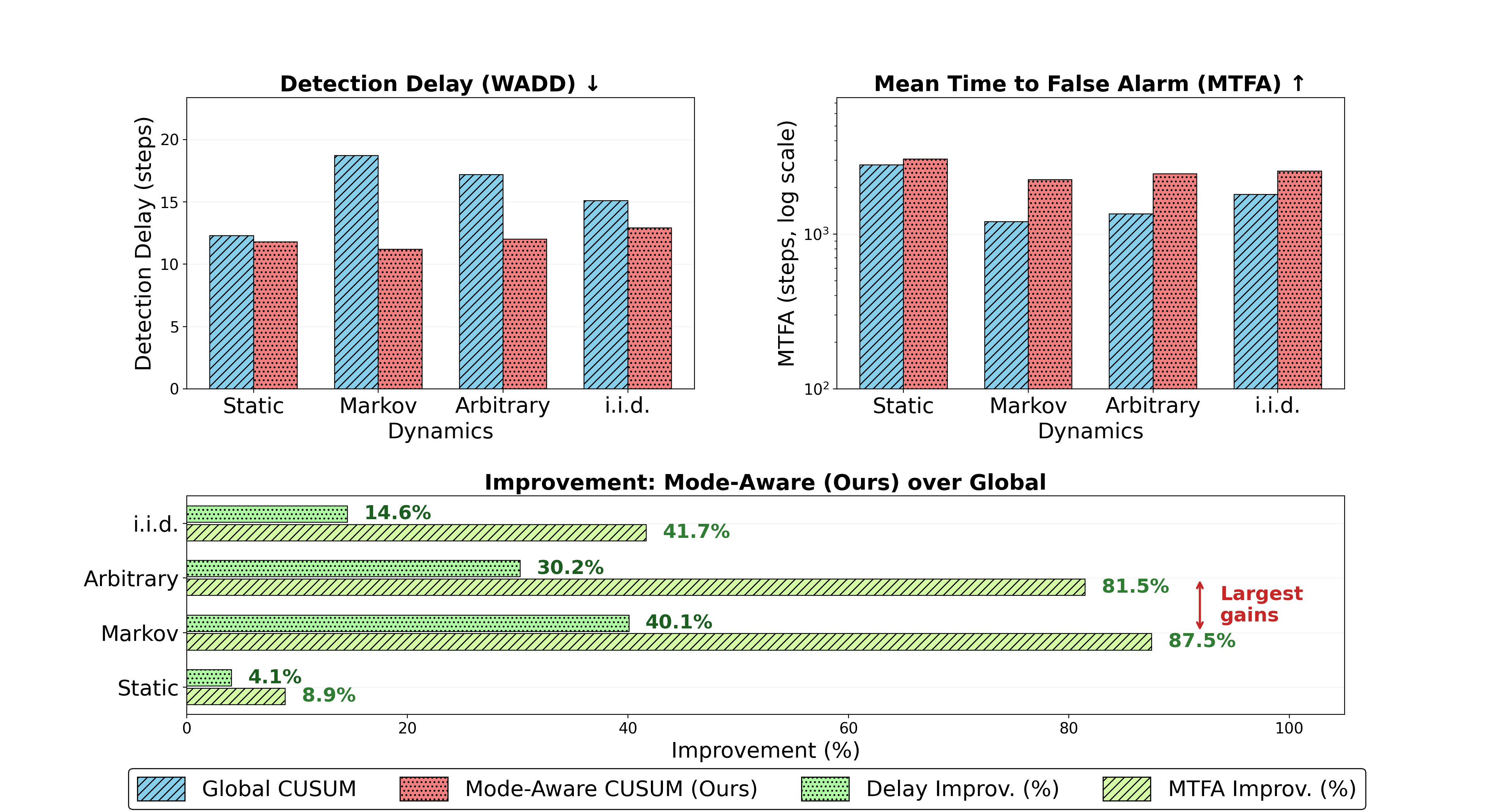}
    \caption{{\bf {Detection performance (delay vs.\ false-alarm rate) of Global CUSUM and Mode-Aware CUSUM under Markov, i.i.d., static, and arbitrary mode-switching models.}} {Synthetic trajectories are generated by GRIP++ across the three datasets in Section~\ref{subsec: experiment setup} (sequence length $T\!=\!3000$ steps, averaged over $N\!=\!5000$ runs per configuration, with both detectors calibrated to a common false-alarm level).}}
    \label{fig:dynamicmode}
\vspace{-10pt}
\end{figure}

\vspace{0.7em}
\noindent\textbf{RQ3: How do the switching dynamics between low- and high-error regimes affect the gains of multi-mode detection?}
The gain of MA-CUSUM scales with how long each regime persists (Table~\ref{tab:dynamics_performance}, Fig.~\ref{fig:dynamicmode}). The \textit{Markov} and \textit{Arbitrary} settings are the closest to real driving. Here MA-CUSUM cuts WADD by $30.2\%$ to $40.1\%$, because when a regime persists long enough, regime-specific scoring more than pays for the lower per-mode update rate. The \textit{Static} case is different. It gains only $4.1\%$, reflecting a small tolerance overhead: near the GMM boundaries the MAP estimator occasionally misassigns the mode, and the stabilizing parameters $\lambda$ and $r_m$ cost a little sensitivity when no switches actually occur. MA-CUSUM still beats the Global baseline in both the \textit{Static} and \textit{i.i.d.} cases. This shows the framework stays robust across a range of non-stationary behaviors and does not suffer negative transfer in simple scenes.

\vspace{0.7em}
\noindent\textbf{RQ4: How robust is multi-mode detection under imperfect post-change knowledge and limited observability?}
Classical CUSUM assumes a fully known post-change distribution $g$, yet real-world robotic deployment rarely affords this luxury. To quantify the impact of such uncertainty, we evaluate three operationally motivated knowledge settings (Section~\ref{sec:knowledge_settings}), ranging from ideal to worst-case.
Table~\ref{tab:robustness} reports results across all three settings. The key finding is that FAR remains stable regardless of post-change uncertainty---even under severe variance misspecification ($\pm50\%\;\sigma$). Detection delay increases monotonically from the oracle baseline (8.2 steps) to the unknown setting (12.4 steps), while AUROC degrades gracefully from 0.95 to 0.895. This reliability arises because the framework anchors its safety guarantees on the pre-change model $f_t$ rather than $g$. Consequently, the system is deployment-ready: it maintains reliable safety bounds while naturally translating better environment knowledge into faster detection.

\begin{table}[t!]
\vspace{0.2em}
\centering
\scriptsize
\setlength{\tabcolsep}{3pt}
\caption{Robustness under varying post-change knowledge.
FAR remains stable;
delay and AUROC degrade gracefully as assumptions weaken.
The unknown-knowledge setting uses $\kappa = 1$
(Section~\ref{sec:knowledge_settings}).}
\label{tab:robustness}
\vspace{0.3em}
\begin{tabular}{llccc}
\toprule
\textbf{Knowledge} & \textbf{Log-likelihood ratio $\ell_t$}
  & \textbf{Delay $\downarrow$}
  & \textbf{FAR $\downarrow$}
  & \textbf{AUROC $\uparrow$} \\
\midrule
Setting~I: Full
  & $\ln \frac{g(\epsilon_t)}{f_{\hat{M}_t}(\epsilon_t)}$
  & 8.2 & 0.034 & 0.95 \\[4pt]
Setting~II: Partial
  & $\ln \frac{\mathcal{N}(\epsilon_t \mid \mu_g,\,\sigma_g^2)}
    {f_{\hat{M}_t}(\epsilon_t)}$
  & 9.1 & 0.036 & 0.93 \\[4pt]
Setting~III: Unknown
  & $\ln \frac{f_{\hat{M}_t}(\epsilon_t - \kappa)}
    {f_{\hat{M}_t}(\epsilon_t)}$
  & 12.4 & 0.035 & 0.895 \\
\bottomrule
\end{tabular}
\end{table}

\begin{table}[t]
\centering
\scriptsize
\caption{Computational cost comparison. Ensemble and MC-Dropout require multiple forward passes; our CUSUM variants add lightweight post-hoc monitoring atop a single pass. $M$: ensemble size, $S$: dropout samples,  $D$: feature dimensions.}
\label{tab:runtime}
\vspace{0.3em}
\resizebox{\columnwidth}{!}{%
\begin{tabular}{lccc}
\toprule
{\bf Method} & {\bf Fwd.}\ {\bf Passes} & {\bf Complexity} & {\bf Time (ms)} \\
\midrule
Deep Ensemble            & $M$ & $O(M \cdot D)$  & 38.7 \\
MC-Dropout               & $S$ & $O(S \cdot D)$  & 29.4 \\
Global CUSUM (base)      & 1   & $O(1)$        & 4.3  \\
\rowcolor{babypurple}
MA-CUSUM (ours)  & 1   & $O(1)$  & 5.1  \\
\bottomrule
\end{tabular}%
}
\end{table}

\subsection{Further Discussions}

\noindent{\bf Computational Efficiency.}
\label{sec:efficiency}
We compare the computational demand of our method against common uncertainty baselines (Table~\ref{tab:runtime}). {Ensemble and MC-Dropout methods require $M$ or $S$ full forward passes, whereas our multi-mode detector operates as a lightweight post-hoc layer on top of a single predictor pass.} Specifically, MA-CUSUM scales {as} $O(1)$, maintaining comparable complexity to
Global CUSUM {while enabling the per-mode statistical decomposition that drives the detection gains reported above}.

\noindent{\bf Update Frequency and Deployment.}
A potential concern is whether partitioning data {across modes} reduces per-mode {sample counts}, thus degrading {detection quality}. Our results demonstrate that {per-mode threshold calibration effectively compensates for the reduced update rate by tightly matching each mode's error distribution}. Quantitatively, even in high-switching scenes (urban intersections),
MA-CUSUM maintains gains of $> 35\%$. {These consistent improvements across diverse traffic environments---from residential corridors to dense urban intersections---indicate that the multi-mode sequential framework is practical for on-board deployment in production autonomous vehicles.}

\section{Conclusion}\label{sec:conclusion}We introduced MA-CUSUM, a multi-mode change-point detector that leverages the modal structure of trajectory prediction errors for adaptive OOD detection. Unlike static or frame-wise baselines, MA-CUSUM maintains context-sensitive thresholds that balance sensitivity across diverse driving scenes. Our results demonstrate that MA-CUSUM consistently outperforms classic baselines and existing UQ methods, achieving lower detection delay and fewer false alarms. This highlights that ``one-size-fits-all'' detection is inherently suboptimal: single thresholds are typically too sensitive in simple scenes or too conservative in complex ones.
Future work will investigate downstream mitigation strategies once OOD is detected, such as safe fallback maneuvers or uncertainty-aware control. We also aim to incorporate
heavy-tailed components (e.g., Student-$t$ mixtures) to better capture rare scenes.

\bibliographystyle{IEEEtran}
\bibliography{IEEEabrv, ref}

\end{document}